%% file: main.tex
\documentclass[11pt]{article}

\usepackage[utf8]{inputenc}
\usepackage[T1]{fontenc}
\usepackage[margin=1in]{geometry}
\usepackage{lmodern}
\usepackage{amsmath,amssymb,amsfonts}
\usepackage{graphicx}
\usepackage{booktabs}
\usepackage{multirow}
\usepackage{array}
\usepackage{xcolor}
\usepackage{microtype}
\usepackage{enumitem}
\usepackage{flafter}
\usepackage{bm}
\usepackage[authoryear,round]{natbib}
\usepackage{hyperref}
\usepackage{cleveref}

\graphicspath{{figures/}}

\title{When Uncertainty Fails to Localize Error: Voxel-Wise Reliability for Sparse-View CT Reconstruction}

\author{Chulin Zhao$^{*}$, Yiran Xu$^{*}$, Gao Yang, and Shu Liu\\
\small $^{*}$Equal contribution}
\date{}

\begin{document}
\maketitle

\begin{abstract}
\input{sections/00_abstract}

\end{abstract}

\begin{figure}[htbp]
  \centering
  \includegraphics[width=\textwidth]{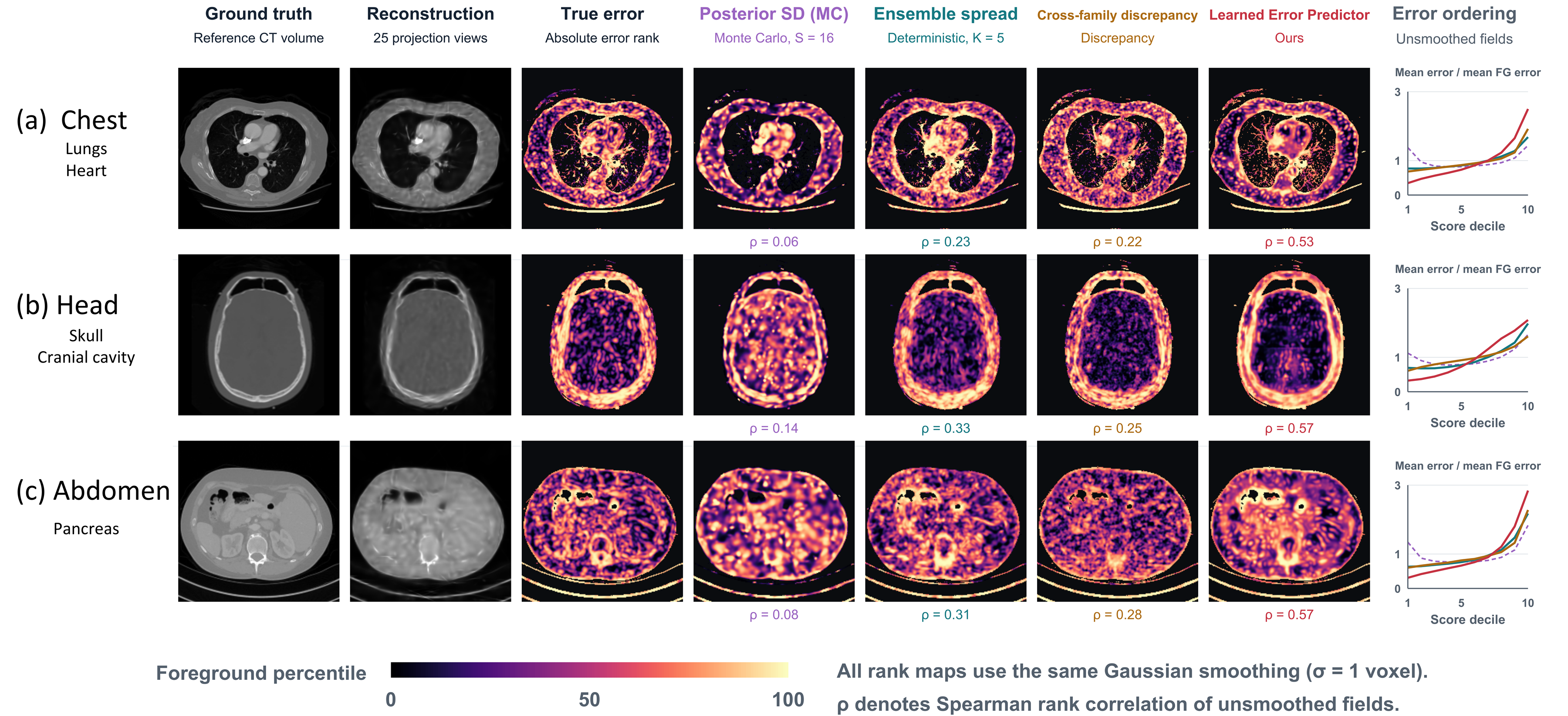}
  \caption{\textbf{Voxel-wise error localization in Sparse View CT.}
  Localization results for multiple methods across three scenes, including ours.
  Predictive dispersion alone shows limited correspondence with error within
  the reconstructed anatomy, while complementary reconstruction-derived
  information enables the learned predictor to recover the spatial ordering
  of voxel-wise reconstruction error more accurately.}
  \label{fig:error-localization-overview}
\end{figure}

\input{sections/01_intro}

\input{sections/02_related}
\input{sections/03_method}
\input{sections/04_experiments}
\input{sections/05_discussion}
\clearpage
\input{sections/07_statements}

\phantomsection\label{marker:refs}%
\bibliographystyle{plainnat}
\bibliography{references}
\clearpage
\appendix
\section*{Appendix}
\input{sections/08_appendix}

\end{document}

%% file: sections/00_abstract.tex
Radiative Gaussian Splatting enables fast and high-quality reconstruction in sparse-view CT, but reliable localization of reconstruction errors within the imaged anatomy remains challenging. It is unclear whether current uncertainty maps can reliably indicate where the reconstruction is wrong. We investigate when uncertainty can serve as a spatial error map and present an error-localization system that predicts voxel-wise reconstruction error by combining predictive uncertainty with complementary reconstruction-derived information. Under fixed Gaussian geometry and independent rectified-Gaussian density variables, an exact closed-form predictive moment is derived in this work to enable direct computation of voxel-wise variance without repeated reconstruction sampling. The resulting uncertainty maps were evaluated over both the whole reconstruction volume and the object foreground. Experiments show that strong whole-volume uncertainty-error correspondence can conceal poor error ranking within the object foreground area. This discrepancy recurs across three tested uncertainty constructions on a 15-scene benchmark. Guided by this analysis, our system combines closed-form predictive dispersion with image, geometry, cross-method discrepancy, and projection-residual cues through a learned error predictor. The resulting system substantially improves voxel-wise error localization within the anatomy, providing a spatially resolved approach to reconstruction reliability assessment in sparse-view CT.

%% file: sections/01_intro.tex
\section{Introduction}
\label{sec:intro}

The usefulness of a CT reconstruction depends on the faithful preservation of local anatomical structures, including small lesions within organs. Improved perceived image quality can coexist with reduced detection of small liver lesions under dose reduction \citep{jensen2022reduced}. In sparse-view CT, standard image-quality metrics can also rank reconstruction methods differently from an observer-based diagnostic task \citep{tivnan2025task}. Reconstruction methods may further fail to recover small structural changes or introduce spurious structures \citep{antun2020instabilities,bhadra2021hallucinations}. These findings motivate spatially resolved assessment of reconstruction reliability: identifying potentially erroneous regions within the imaged anatomy, alongside reconstructing the volume itself.

Uncertainty maps~\citep{edupuganti2021uncertainty,hu2021learning} provide a natural mechanism for spatially characterizing reconstruction reliability. Existing approaches commonly generate multiple candidate reconstructions through posterior sampling~\citep{edupuganti2021uncertainty,luo2023bayesian}, independent model training~\citep{lakshminarayanan2017simple}, or deliberate perturbations~\citep{goli2024bayesrays,wu2026active,vanlohuizen2026aleatoric}. At each voxel, they quantify uncertainty by measuring how the predicted
attenuation coefficients change among different reconstructions, typically
using the variance or other measures of disagreement across predictions to
encode inter-reconstruction dispersion~\citep{lakshminarayanan2017simple,kuestner2024predictive}..

However, an uncertainty map may exhibit strong correspondence with realized reconstruction error when evaluated over the whole volume, without reliably ranking voxel-wise errors within the same anatomical structure~\citep{vanlohuizen2026aleatoric}. In sparse-view CT, whether uncertainty maps produced by existing methods can serve as spatial proxies for realized voxel-wise reconstruction error within the target anatomy remains insufficiently investigated.

In this work, we first find a way to output uncertainty for Radiative Gaussian sparse-view CT reconstruction. By exploiting the linear attenuation structure of X-ray formation under fixed Gaussian geometry, we derive closed-form predictive moments that enable direct computation of voxel-wise predictive variance. However, obtaining accurate predictive variance alone is insufficient for reliable error localization within the anatomy. Further analysis shows that the information contained in predictive dispersion cannot fully characterize these local reconstruction errors. Based on this analysis, we introduce complementary image, geometry, cross-method discrepancy, and projection-residual cues. A learned predictor is trained, mapping all the information to voxel-wise reconstruction error. Figure~\ref{fig:error-localization-overview} visualizes the error-localization result, showing how complementary reconstruction-derived information enables substantially improved correspondence with true voxel-wise error compared with predictive dispersion alone.

In summary, our contributions are threefold: (1) exact closed-form predictive moments for Radiative Gaussian reconstruction; and (2) a systematic analysis of why uncertainty maps that reflect reconstruction error over the whole volume may fail to localize errors within the anatomy; and (3) an error-localization system that combines predictive dispersion with reconstruction-derived cues to predict voxel-wise reconstruction error.

%% file: sections/02_related.tex
\section{Related Work}

\subsection{Radiative Gaussian Reconstruction for Sparse View CT}
Continuous representations enable three-dimensional attenuation fields to be optimized directly from sparse CT projections. NAF, for example, parameterized attenuation as a neural function of spatial coordinates~\citep{zha2022naf}. Building on 3D Gaussian Splatting~\citep{kerbl20233dgs}, X-Gaussian adapted explicit Gaussian representations and differentiable rasterization to X-ray novel-view synthesis~\citep{cai2024xgaussian}. R$^2$-Gaussian corrected an integration bias in Gaussian projection and introduced differentiable voxelization for direct tomographic reconstruction~\citep{zha2024r2gaussian}. These advances have established efficient representations and computational tools for reconstruction. Our work builds on these foundations but focuses on the reliability of the resulting volumes.

\subsection{Uncertainty Estimation and Analytic Computation}

Predictive uncertainty is commonly estimated from multiple prediction samples, obtained through stochastic forward passes with Monte Carlo dropout~\citep{gal2016dropout}, independently trained deep ensembles~\citep{lakshminarayanan2017simple}, or perturbations that produce multiple reconstructions~\citep{goli2024bayesrays,wu2026active}. These approaches estimate predictive uncertainty from variation across samples and therefore require repeated predictions or reconstructions.

Sampling-free approaches instead seek to obtain predictive uncertainty without repeatedly generating prediction samples. \citet{postels2019sampling} approximated predictive moments by propagating uncertainty through neural-network operations, thereby avoiding Monte Carlo sampling. VarSplat explicitly modeled per-Gaussian appearance variance and computed pixel-wise uncertainty in a single rasterization pass~\citep{tran2026varsplat}. The former relies on approximate moment propagation, while the latter uses learned variance. In contrast, we exploit the mathematical structure of the Radiative Gaussian CT formation process to derive exact closed-form predictive moments, enabling direct computation of voxel-wise predictive variance without repeated sampling.

\subsection{Evaluating Spatial Uncertainty in Reconstruction}

Spatial uncertainty maps are commonly evaluated by comparison with realized reconstruction error. At the whole-image or whole-volume scale, correlation metrics can measure the correspondence between uncertainty and voxel-wise error, while ranking-based metrics such as sparsification assess whether locations with higher uncertainty also tend to exhibit larger reconstruction errors~\citep{ilg2018uncertainty}. Region-conditioned evaluation instead restricts the evaluation domain to a target anatomical region and compares uncertainty and reconstruction error only among voxels within that region. Changing the evaluation domain can substantially alter the observed uncertainty--error relationship. For example, in accelerated prostate MRI, \citet{vanlohuizen2026aleatoric} reported markedly weaker uncertainty--error correlations within the prostate than across the whole slice. Recomputing correlation or error-ranking metrics within the target region therefore provides a more direct assessment of whether an uncertainty map can localize reconstruction errors within the anatomy.

\subsection{Learned Reconstruction Error Prediction}
\citet{hu2021learning} distinguished predictive uncertainty from reconstruction error and trained a separate error predictor for MRI reconstruction. Building on this two-stage perspective, we ask which observable cues improve foreground error ranking when whole-volume uncertainty evaluation is misleading. In sparse-view Radiative Gaussian CT, we diagnose this mismatch across three uncertainty constructions and combine closed-form predictive dispersion with image, geometry, cross-method discrepancy, and projection-residual cues. Matched feature and supervision controls under scene-level holdout test the contribution of these cues to foreground error localization.

%% file: sections/03_method.tex
\section{Method}
\label{sec:method}

\subsection{Problem Formulation and Evaluation Target}
\label{sec:problem}

Given sparse-view CT measurements $y$ after a negative-log transform
and a known projection operator $A$, our goal is to reconstruct a
three-dimensional attenuation volume $\widehat{V}$ and produce a
voxel-wise error score $s(\mathbf{x})$. Here $\mathbf{x}$ denotes a
voxel location, and larger scores should indicate larger errors in
the accompanying reconstruction. Neither output requires a reference
volume at inference.

\begin{figure*}[t]
    \centering
    \includegraphics[width=\textwidth]{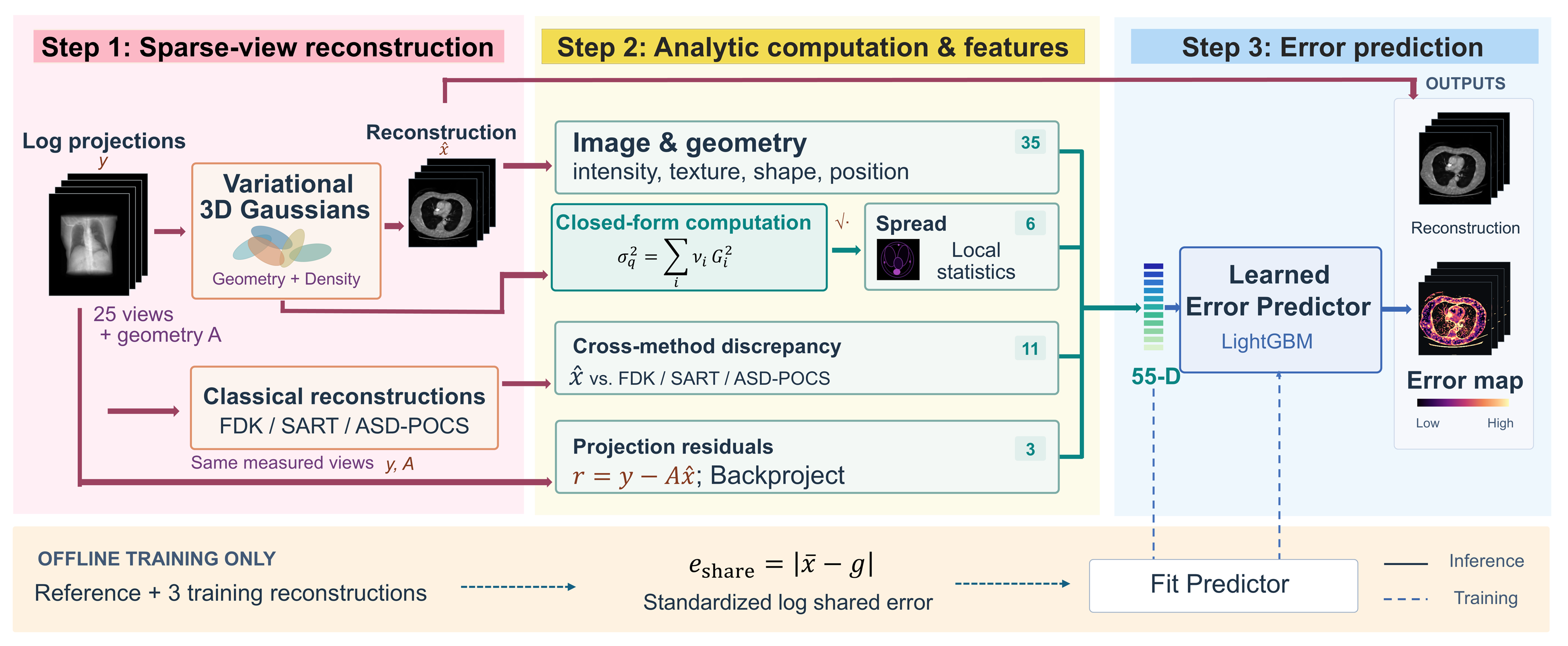}
    \caption{\textbf{Overall workflow.} 
    Sparse-view log projections and acquisition geometry are used to obtain a Radiative Gaussian reconstruction and auxiliary classical reconstructions. Closed-form predictive dispersion is combined with image-and-geometry features, cross-method discrepancy, and projection-residual cues to form a 55-dimensional voxel representation, which is mapped to an error score $s(\mathbf{x})$ by a learned error predictor. Inference (solid arrows) requires no reference volume.}
    \label{fig:error-localization-framework}
\end{figure*}

For a reference volume $V^{*}$, the reconstruction error is defined as
\begin{equation}
  \label{eq:error-target}
  e(\mathbf{x}) = \bigl|\widehat{V}(\mathbf{x}) - V^{*}(\mathbf{x})\bigr|.
\end{equation}
Reference volumes are used only for offline supervision and evaluation.
Each score map is evaluated against the error of the corresponding reconstruction over both the whole volume and the
object foreground. Spearman correlation measures agreement between
the voxel rankings induced by $s$ and $e$, with higher values indicating
better agreement. The area under the sparsification error curve (AUSE)
measures how effectively the score identifies high-error voxels.
We progressively remove voxels with the highest scores and compute
the root mean squared error (RMSE) of the remaining voxels. AUSE
integrates the gap between this curve and an oracle curve obtained
by removing voxels with the largest true errors, normalized by the
initial RMSE of the evaluated region. Lower AUSE indicates better
error localization. An overview of the proposed error-localization system is shown in Figure~\ref{fig:error-localization-framework}.

\subsection{Density Modeling and Closed-Form Uncertainty}
\label{sec:closed-form}

Represent the attenuation volume using $N$ Gaussian primitives:
\begin{equation}
  \label{eq:gaussian-kernel}
  g_i(\mathbf{x}) = \exp\!\left[
    -\frac{1}{2}(\mathbf{x}-\mathbf{p}_i)^{\top}
    \Sigma_i^{-1}(\mathbf{x}-\mathbf{p}_i)\right],
  \qquad
  V(\mathbf{x}) = \sum_{i=1}^{N} a_i g_i(\mathbf{x}).
\end{equation}
Here $g_i$ is the spatial kernel of primitive $i$, $\mathbf{p}_i$ is its center,
$\Sigma_i$ is the positive-definite spatial covariance controlling
shape and orientation, and $a_i$ is the density weight. The superscript
$\top$ denotes transpose. Deterministic density weights are replaced
with independent Gaussian latent variables and rectify them to preserve
nonnegative attenuation:
\begin{equation}
  \label{eq:stochastic-density}
  z_i \sim \mathcal{N}(m_i,\tau_i^2),
  \qquad a_i = \max(0,z_i).
\end{equation}
The normal distribution has mean $m_i$ and standard deviation $\tau_i$;
both are constrained to be positive during training. Let $q$ denote
the resulting joint distribution of the rectified weights $a_i$.
The deterministic reconstruction uses $m_i$ directly:
\begin{equation}
  \label{eq:deployed-reconstruction}
  \widehat{V}(\mathbf{x}) = \sum_{i=1}^{N} m_i g_i(\mathbf{x}).
\end{equation}

To learn the density variances, we sample
$z_i=m_i+\tau_i\epsilon_i$, where the $\epsilon_i$ are independent
standard normal draws, and minimize the training objective
\begin{equation}
  \label{eq:density-training}
  \mathcal{L}
  = \mathrm{E}_{q}\!\left[
      \mathcal{L}_{\mathrm{proj}}(AV,y)
      + \lambda_{\mathrm{TV}}\mathrm{TV}(V)\right]
    + \lambda_{\mathrm{reg}}\mathcal{R},
\end{equation}
\begin{equation}
  \label{eq:density-scale-regularizer}
  \mathcal{R} = \frac{1}{2N}\sum_{i=1}^{N}
  \left[
    \frac{\tau_i^2}{\tau_0^2}
    - \log\!\left(\frac{\tau_i^2}{\tau_0^2}\right) - 1
  \right].
\end{equation}
$\mathrm{E}_{q}$ averages over the random density weights.
The projection loss $\mathcal{L}_{\mathrm{proj}}$ combines mean absolute
error with a weighted structural-similarity loss between rendered
projections $AV$ and measurements $y$. The total-variation penalty
$\mathrm{TV}$ promotes spatial smoothness, while $\mathcal{R}$ encourages
the density scales toward a fixed reference scale $\tau_0>0$.
The nonnegative weights $\lambda_{\mathrm{TV}}$ and
$\lambda_{\mathrm{reg}}$ control their strengths.
One density sample is used per iteration, with sampled views and volume
patches for projection and spatial losses. Backpropagation updates
$m_i$, $\tau_i$, and the Gaussian geometry jointly.

After training, we fix the geometry and compute predictive moments
analytically. Define $\alpha_i=m_i/\tau_i$, and let $\Phi$ and $\phi$
be the cumulative distribution and density functions of a standard
normal variable. Integrating the positive part of each Gaussian gives
the rectified density mean $\mu_i^{+}$ and variance $v_i^{+}$:
\begin{align}
  \label{eq:rectified-mean}
  \mu_i^{+} &= \mathrm{E}_{q}[a_i]
    = m_i\Phi(\alpha_i)+\tau_i\phi(\alpha_i), \\
  \label{eq:rectified-variance}
  v_i^{+} &= \mathrm{Var}_{q}(a_i)
    = (m_i^2+\tau_i^2)\Phi(\alpha_i)
      +m_i\tau_i\phi(\alpha_i)-(\mu_i^{+})^2.
\end{align}
Here $\mathrm{Var}_{q}$ denotes variance under $q$. Linearity and independence give the closed-form predictive moments
\begin{equation}
\label{eq}
\mu_q(\mathbf{x})=\sum_{i=1}^{N}g_i(\mathbf{x})\mu_i^{+},
\qquad
\sigma_q^2(\mathbf{x})=\sum_{i=1}^{N}g_i(\mathbf{x})^2v_i^{+}.
\end{equation}
Rectification means that $\mu_q$ generally differs from the returned reconstruction $\widehat{V}$. Since $g_i^2$ is a Gaussian kernel with covariance $\Sigma_i/2$, the variance map can be computed by voxelization with weights $v_i^{+}$, without repeated sampling. These moments are exact for the stated linear model with fixed geometry and independent densities.

We use $\sigma_q^2(\mathbf{x})$ as the uncertainty map; its square root
$\sigma_q(\mathbf{x})$ is the predictive standard deviation.
Either can serve as the score $s(\mathbf{x})$ in spatial ranking,
since they induce the same ordering. They quantify model dispersion;
their ability to localize the reconstruction error $e(\mathbf{x})$
is assessed using the metrics in the preceding subsection.

\subsection{Regional Diagnosis of Error Localization}
\label{sec:diagnosis-framework}
Let $\Omega$ denote all voxels and
$V_{\max}^{*}=\max_{\mathbf{x}\in\Omega}V^{*}(\mathbf{x})$ the peak
reference attenuation. For offline evaluation, we partition $\Omega$ into
background $\mathcal{B}=\{V^{*}\leq0.01V_{\max}^{*}\}$, transition
$\mathcal{T}=\{0.01V_{\max}^{*}<V^{*}\leq0.05V_{\max}^{*}\}$, and
foreground $\mathcal{F}=\{V^{*}>0.05V_{\max}^{*}\}$. The transition region contains intermediate
attenuation values and need not coincide with a geometric boundary.
We compare whole-volume and foreground Spearman correlation and AUSE,
holding the reconstruction $\widehat{V}$, its error $e(\mathbf{x})$,
and the score map $s(\mathbf{x})$ fixed. This tests whether good global
ranking also reflects useful error localization within the object.

To quantify the contribution of regional differences, raw score values are analyzed, using $s(\mathbf{x})=\sigma_q(\mathbf{x})$ for the
closed-form model. Let $G$ be the region label of a uniformly sampled
voxel. The law of total covariance gives
\begin{equation}
\label{eq:cov-decomp}
\mathrm{Cov}(s,e)
=
\underbrace{\mathrm{E}_{G}\!\left[\mathrm{Cov}(s,e\mid G)\right]}_{C_{\mathrm{within}}}
+
\underbrace{\mathrm{Cov}_{G}\!\left(\mathrm{E}[s\mid G],\mathrm{E}[e\mid G]\right)}_{C_{\mathrm{between}}}.
\end{equation}
Here $\mathrm{E}$ and $\mathrm{Cov}$ denote spatial means and covariances;
conditioning on $G$ restricts them to a region, and the outer operations
weight regions by their voxel proportions. The within-region term
$C_{\mathrm{within}}$ measures association among voxels in the same region,
while $C_{\mathrm{between}}$ captures association between regional means.
A large positive between-region contribution indicates that coarse
regional contrasts contribute strongly to the global association.
This identity decomposes raw covariance, not Spearman correlation.

We further examine errors shared across reconstruction members.
For $K$ member volumes $V_k$, define their mean
$\bar{V}(\mathbf{x})=K^{-1}\sum_{k=1}^{K}V_k(\mathbf{x})$
and their individual errors
$e_k(\mathbf{x})=|V_k(\mathbf{x})-V^{*}(\mathbf{x})|$.
Their mean squared error satisfies
\begin{equation}
\label{eq:mse-decomp}
\frac{1}{K}\sum_{k=1}^{K} e_k(\mathbf{x})^2
=
\bigl(\bar{V}(\mathbf{x})-V^{*}(\mathbf{x})\bigr)^2
+
\frac{1}{K}\sum_{k=1}^{K}
\bigl(V_k(\mathbf{x})-\bar{V}(\mathbf{x})\bigr)^2.
\end{equation}
The first term is the error of the member mean, termed finite-member
shared squared error; the second is variation around that mean.
The regional shared fraction is the sum of the first term divided by
the sum of the left-hand side over the region.
We compare the Spearman correlations of $\sigma_q$ with shared absolute
error $|\bar{V}-V^{*}|$ and with the square root of the second term.
Stronger association with member variation suggests that dispersion
tracks disagreement more closely than shared error. A large shared
fraction alone does not establish poor error ranking.
Independently trained members and samples from a fixed density model
are analyzed separately.

Finally, to test the effect of the difference between predictive mean
$\mu_q$ and returned reconstruction $\widehat{V}$, we evaluate the score
\begin{equation}
  \label{eq:center-corrected-score}
  s_{\mathrm{center}}(\mathbf{x})
  = \sqrt{\sigma_q^2(\mathbf{x})
    +(\mu_q(\mathbf{x})-\widehat{V}(\mathbf{x}))^2}.
\end{equation}
This is the model's root mean squared distance from $\widehat{V}$.
Comparing it with $\sigma_q$ against the same error $e(\mathbf{x})$
tests whether accounting for this mean offset improves localization.

\subsection{Learning to Predict Reconstruction Errors}
\label{sec:learned-predictor}

We complement predictive dispersion with reconstruction-derived cues.
For each voxel, a 55-dimensional feature vector $\mathbf{h}(\mathbf{x})$
contains four groups: image and geometry descriptors (IMG, 35),
including intensity, local statistics, gradients, and boundary distances;
dispersion descriptors derived from $\sigma_q$ (SIG, 6);
discrepancies with three classical reconstructions, FDK, SART, and
ASD-POCS, from the same measurements (DIS, 11); and backprojected
descriptors of the projection residual $r=A\widehat{V}-y$ (RES, 3).
A gradient-boosted tree regressor implemented with LightGBM produces
\begin{equation}
  \label{eq:learned-error-score}
  s(\mathbf{x})=f_{\theta}\bigl(\mathbf{h}(\mathbf{x})\bigr),
\end{equation}
where $f_{\theta}$ is the regressor and $\theta$ its learned parameters.

For supervision, we use the shared absolute error
$e_{\mathrm{shared}}(\mathbf{x})=|\bar{V}(\mathbf{x})-V^{*}(\mathbf{x})|$,
where $\bar{V}$ averages three independently trained reconstructions
of each training scene. The regression target is
\begin{equation}
  \label{eq:predictor-training-target}
  t(\mathbf{x})=
  \frac{\log\!\bigl(e_{\mathrm{shared}}(\mathbf{x})+\delta\bigr)
        -\mu_{\log}}{\sigma_{\log}}.
\end{equation}
Here $\delta>0$ stabilizes the logarithm. $\mu_{\log}$ and
$\sigma_{\log}>0$ are the mean and numerically stabilized standard
deviation of the log errors, computed separately over each training
scene's foreground pool. With reconstruction parameters fixed, we
fit $f_{\theta}$ by squared-error regression to $t$.
Replacing $e_{\mathrm{shared}}(\mathbf{x})$ with the individual
reconstruction error $e(\mathbf{x})$ provides a supervision control.

Leave-one-scene-out evaluation is applied: each fold trains on the remaining
scenes and evaluates all three reconstruction seeds of the held-out
scene. Target transformations use training scenes only. Ablations
share the sampled voxels and regressor capacity. Predictions are
assessed against each returned reconstruction's error $e(\mathbf{x})$
using the regional Spearman correlation and AUSE defined above.
Reference volumes provide supervision and offline evaluation masks;
inference feature values use only measurements and reconstructions. In the separate human-anatomy study, a 3D U-Net receives eight
image-derived channels and two coordinate channels. We average its
within-scene voxel ranks equally with those of an anatomy-trained
tree model. Training and test subjects are disjoint.

%
%
%
%

%% file: sections/04_experiments.tex
\section{Experiments}
\label{sec:experiments}

\subsection{Validating Closed-Form Predictive Moments}
\label{sec:exp-closed-form}

We compare closed-form (CF) predictive moments $\mu_q$ and $\sigma_q^2$
with Monte Carlo (MC) estimates on chest, foot, and beetle, using one
frozen 25-view model per scene and $256^3$ voxel grids. Both methods
use the same geometry and rectified density distribution $q$.
Five independent sampling streams are evaluated at
$S\in\{16,32,64,128,256,512\}$ sampled volumes; MC variance uses the
$S-1$ denominator. Agreement is evaluated in the foreground
$\mathcal{F}$. Relative $L_2$ is the norm of the MC--CF difference
divided by the norm of the corresponding CF map.

\begin{table}[htbp]
  \caption{CF--MC agreement and readout efficiency at representative
  sample budgets. Mean and variance
  relative $L_2$ differences compare MC estimates with $\mu_q$ and
  $\sigma_q^2$, respectively, and are reported in percent.
  Pearson is the correlation between variance-map values.
  Agreement entries are medians over five sampling seeds within each
  scene, followed by the median across three scenes.}
  \label{tab:closed-form-mc}
  \begin{center}
  \begin{tabular}{@{}rcccc@{}}
    \hline
    MC samples $S$ & \shortstack{Mean rel. $L_2$\\(\%)}
      & \shortstack{Variance rel. $L_2$\\(\%)}
      & \shortstack{Variance\\Pearson}
      & \shortstack{Readout\\speedup} \\
    \hline
    16  & 1.60 & 42.86 & 0.9030 & $8.0\times$ \\
    128 & 0.57 & 15.44 & 0.9870 & $62.1\times$ \\
    512 & 0.28 & 7.66  & 0.9965 & $247.7\times$ \\
    \hline
  \end{tabular}
  \end{center}
\end{table}

Table~\ref{tab:closed-form-mc} shows decreasing MC--CF differences
as the sample count increases. In a separate float64 check, we fix
spatial support and disable density-dependent response pruning at
1,024 coordinates per scene. Across the 512 foreground coordinates
per scene, the five-seed median variance relative $L_2$ at $S=512$
is 5.41\%--5.89\%.

We time full-volume predictive mean and
variance readout, including MC sampling and accumulation.
Each configuration uses two warm-up runs and five timed repetitions,
with GPU synchronization at the timing boundaries; training, model
loading, and disk I/O are excluded. CF takes 0.10--1.51\,s, compared
with 29.03--374.32\,s for MC with $S=512$, giving scene-wise speedups
of 247--278$\times$. Including transfer to host memory, the speedups
remain 166--228$\times$.

\subsection{Foreground Error Localization and Shared Errors}
\label{sec:exp-diagnosis}

Regional evaluation uses the 15-scene R$^2$-Gaussian benchmark with
25 views, $512{\times}512$ projections, and $256^3$ volumes.
Closed-form uncertainty is compared
with a log-normal density control
and disagreement across independently trained deterministic reconstructions.
The log-normal control uses exponentiated Gaussian latent variables as strictly positive density weights, testing whether the foreground-ranking deficit persists under an alternative density parameterization.

\begin{figure*}[t]
    \centering
    \includegraphics[width=\textwidth]{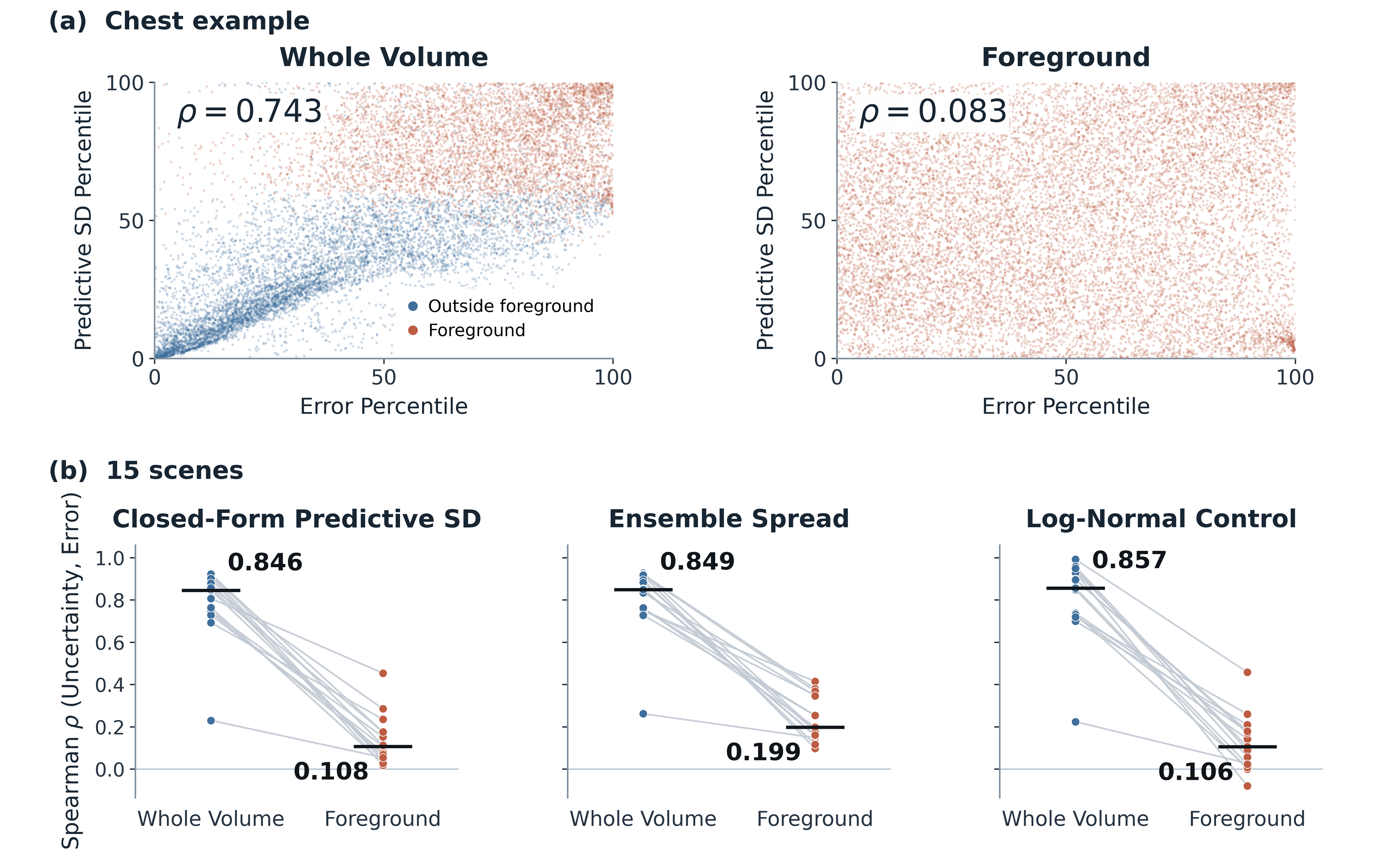}
    \caption{\textbf{Uncertainty--error correspondence deteriorates within
    the reconstruction foreground.}
    (a) Representative chest case using the closed-form predictive standard
    deviation of the rectified-Gaussian model. Each point represents a voxel,
    with reconstruction error and predictive standard deviation shown as
    percentiles.
    (b) Whole-volume and foreground Spearman correlations across the 15 test
    scenes for closed-form predictive standard deviation, deterministic-
    ensemble spread, and the log-normal density control.}
    \label{fig:regional-failure}
\end{figure*}

Figure~\ref{fig:regional-failure} shows that strong whole-volume uncertainty--error correspondence deteriorates sharply when evaluation is restricted to the foreground. This decline is consistent across the 15 test scenes and all three uncertainty constructions (Table~\ref{tab:regional-ranking}). The reconstruction and score map are held fixed when changing the evaluation region.

\begin{table}[htbp]
  \caption{Regional Spearman correlation $\rho$ between the uncertainty
  score $s(\mathbf{x})$ and reconstruction error $e(\mathbf{x})$;
  higher is better. Whole-volume and foreground refer to $\Omega$ and
  $\mathcal{F}$ in Section~\ref{sec:diagnosis-framework}.
  Scores are predictive standard deviations for the two density models
  and standard deviation across five deterministic members for the
  ensemble. Ensemble error is measured from the member mean.
  Entries are medians across 15 scenes, after averaging three
  reconstruction seeds for each density model.}
  \label{tab:regional-ranking}
  \begin{center}
  \begin{tabular}{@{}lcc@{}}
    \hline
    Uncertainty construction & Whole-volume $\rho$ & Foreground $\rho$ \\
    \hline
    Rectified-Gaussian density & 0.846 & 0.108 \\
    Deterministic ensemble    & 0.849 & 0.199 \\
    Log-normal density       & 0.857 & 0.106 \\
    \hline
  \end{tabular}
  \end{center}
\end{table}

For the rectified-Gaussian model,
the between-region term in Eq.~\eqref{eq:cov-decomp} accounts for a
median 81.4\% of the raw covariance between $\sigma_q$ and $e$ across
15 seed-0 reconstructions. This indicates a strong contribution from
coarse regional differences; it does not decompose Spearman correlation.
The decline persists at foreground thresholds of 1\%, 2\%, 5\%, and 10\%
for the rectified-Gaussian model and ensemble.

A separate control evaluates $\sigma_q$ against the predictive mean's
own absolute error, using foreground masks defined from fixed
reconstructions at 5\% of their positive-intensity 99.99th percentile.
After averaging three seeds within each scene,
mean correlation across the benchmark falls from 0.809 over the
whole volume to 0.146 in this foreground, and remains 0.197 more than
five voxels inside its boundary. Thus, reference-defined masks and
the outermost boundary do not fully explain the deficit.

The finite-member decomposition in Eq.~\eqref{eq:mse-decomp} identifies
a possible explanation. Shared squared error accounts for median
foreground fractions of 0.890 for five deterministic members and
0.935 for three independently trained variational reconstructions.
In a separate analysis of these three reconstruction seeds, seed-0
$\sigma_q$ correlates more strongly with variation across members
than with their shared absolute error $|\bar{V}-V^{*}|$
(median foreground Spearman: 0.427 versus 0.101).
A common spatial error can shift all members together without changing
their variance. Together, these observations suggest that uncertainty
captures variation across reconstructions more readily than errors
they repeatedly share; a large shared fraction alone would not establish
poor localization. Correcting the predictive-mean offset using
$s_{\mathrm{center}}$ improves mean foreground correlation by only
0.0014 (95\% scene-bootstrap interval: [0.0001, 0.0028]).

Error-localization performance also depends on the reconstruction method.
On eight human CT volumes with simulated projections, the standard
deviation across diffusion posterior samples achieves median foreground
Spearman correlation 0.376 with the error of their mean reconstruction.
When the same diffusion uncertainty maps are instead evaluated against
the Gaussian reconstruction error, correlation decreases to 0.175.

\subsection{Diagnosis-Driven Error Prediction}
\label{sec:exp-repair}

The 55-feature learned error predictor, designed based on our analysis, is evaluated on the 15-scene,
25-view benchmark using leave-one-scene-out evaluation.
Each fold uses 80,000 foreground voxels from reconstruction seed 0
per training scene,
1,200 boosting rounds, at most 255 leaves per tree, and learning
rate 0.04. Reconstruction parameters remain fixed. IMG/Individual serves as a supervised image-and-geometry baseline with matched training voxels, regressor settings, and fitting budget. Full/Individual adds SIG, DIS, and RES under the same supervision, while Full/Shared additionally changes the training target.

\begin{table}[htbp]
  \caption{Foreground error localization against the same
  $e(\mathbf{x})$ on common evaluation voxels.
  $\rho$ denotes Spearman correlation (higher is better), and AUSE
  is normalized RMSE AUSE (lower is better).
  Full combines IMG, SIG, DIS, and RES from
  Section~\ref{sec:learned-predictor}; parenthesized numbers are feature
  counts.}
  \label{tab:matched-learner}
  \begin{center}
  \begin{tabular}{@{}llcc@{}}
    \hline
    Features or score & Supervision & Foreground $\rho$ & AUSE \\
    \hline
    Full (55, ours)      & Shared     & 0.399 & 0.304 \\
    Full (55)            & Individual & 0.398 & 0.310 \\
    IMG+DIS+RES (49)      & Shared     & 0.391 & 0.316 \\
    IMG (35)             & Individual & 0.331 & 0.376 \\
    Raw dispersion       & None       & 0.136 & 0.542 \\
    \hline
  \end{tabular}
  \end{center}
\end{table}

Under individual-error supervision, jointly adding SIG, DIS, and RES improves both mean metrics over IMG alone (Table~\ref{tab:matched-learner}). The Full/Shared configuration improves both metrics over IMG/Individual in all 15 scenes for each of the three learner seeds. Intervals use 20,000 scene resamples; component comparisons are unadjusted.

Feature ablations with one learner seed and fixed shared-error
supervision show a mean foreground Spearman gain of 0.0269 from
adding DIS to IMG. RES adds 0.0327 to IMG and 0.0233 to IMG+DIS.
Cross-method discrepancy features (DIS) compare reconstructions from
different methods, whose
error patterns can differ even when repeated reconstruction members of one method agree.
Projection-residual features (RES) provide evidence of disagreement
with the measured projections,
while image and geometry features (IMG) describe local structure.
These cues can therefore help identify errors that persist across
reconstruction seeds, consistent with the diagnosis in
Section~\ref{sec:exp-diagnosis}. This is a mechanism interpretation,
not a causal demonstration.

Across three learner seeds, dispersion features (SIG) provide a smaller
conditional Spearman gain of 0.0086 [0.0010, 0.0147].
With all 55 features fixed, shared-error supervision yields only
0.0014 [-0.0060, 0.0082] over individual-error supervision, although
AUSE improves slightly. Thus, the observed benefit is better supported
by complementary inputs than by a distinct advantage of shared-error
supervision. The lower AUSE indicates a smaller integrated gap between score-based and oracle sparsification curves. This provides an offline measure of how effectively the score prioritizes high-error voxels while keeping the reconstructed volume fixed. Additional transfer evaluations using earlier predictor configurations
are provided in the appendix.

Human anatomy is evaluated using the fixed, equal-weight rank fusion
of the tree predictor and 3D U-Net in
Section~\ref{sec:learned-predictor}.
Anatomy-augmented training adds ten MSD subjects to the benchmark;
the original test cohort contains four other MSD subjects and four
LUNA16 subjects. All human volumes use simulated 25-view projections.
The anatomy-augmented fusion improves over benchmark-only fusion
in all eight subjects at reconstruction seed 0, with a median paired
gain of 0.0561 [0.0340, 0.0893].
Table~\ref{tab:anatomy} also reports reconstruction-seed averaging and
a retrospective evaluation on eight additional subjects using frozen
predictor checkpoints.

\begin{table}[htbp]
  \caption{Median foreground Spearman correlation across human subjects.
  Anatomy and Benchmark denote the rank-fusion predictors trained with
  and without the ten additional human training subjects, respectively;
  Dispersion denotes raw $\sigma_q$. The final row uses eight additional subjects and frozen predictors.}
  \label{tab:anatomy}
  \begin{center}
  \begin{tabular}{@{}lccc@{}}
    \hline
    Evaluation & Anatomy & Benchmark & Dispersion \\
    \hline
    Original 8, seed 0       & 0.566 & 0.483 & 0.068 \\
    Same 8, three-seed mean  & 0.544 & 0.464 & 0.067 \\
    Additional 8, seed 0     & 0.537 & 0.468 & 0.045 \\
    \hline
  \end{tabular}
  \end{center}
\end{table}

%
%

%% file: sections/05_discussion.tex
\section{Discussion and Conclusion}
\label{sec:discussion}

We derive exact predictive moments for our linear attenuation model
under fixed Gaussian geometry and independent rectified-Gaussian
densities. Our experiments show that strong whole-volume
uncertainty--error correspondence can conceal poor foreground error
ranking. Combining predictive dispersion with complementary
reconstruction-derived cues improves voxel-wise error localization
in Radiative Gaussian sparse-view CT.

Anatomy-specific training further improves localization, possibly by
learning domain-specific error patterns. However, our human experiments
use simulated projections; generalization to real clinical acquisitions,
scanners, and protocols remains untested. The scores could prioritize
inspection of high-scoring locations without a reference volume at
inference. Our results establish improved foreground error ranking;
benefits for reader decisions, lesion detection, and downstream
correction require further evaluation.

%% file: sections/07_statements.tex
\subsection*{AI use statement}
In this work, generative AI tools are used to polish writing, search for related work and check references. Coding agents are used for code implementation and experiment execution. The authors take responsibility for the final content of this work, including its implementation, experimental results, references, and scientific claims.

\subsection*{Ethics statement}
This study uses existing publicly available CT volumes and specimen scans, with no new participant recruitment or scanning. All human CT experiments use simulated sparse-view projections. The proposed scores are evaluated for reconstruction-error ranking and localization; they have not been validated for clinical decision-making, lesion detection, or calibrated error estimation. Their use to support clinical interpretation would require further validation under real acquisition conditions and across patient populations, scanners, and imaging protocols.

\subsection*{Reproducibility statement}
Section~\ref{sec:closed-form} specifies the density model, assumptions, and closed-form predictive moments. Sections~\ref{sec:problem} and~\ref{sec:diagnosis-framework} define the evaluation metrics and regions, while Section~\ref{sec:learned-predictor} describes the predictor inputs, supervision, and scene-level evaluation splits. Section~\ref{sec:exp-closed-form} details the Monte Carlo comparison and timing protocol. Sections~\ref{sec:exp-diagnosis} and~\ref{sec:exp-repair} report the benchmark settings, predictor hyperparameters, seed aggregation, and scene-bootstrap procedure.

%% file: sections/08_appendix.tex

\section{Closed-Form Moments and Numerical Checks}
\label{app:cf-validation}

\paragraph{Derivation and scope.}
Using $z_i=m_i+\tau_i\epsilon_i$ with
$\epsilon_i\sim\mathcal{N}(0,1)$ as in Section~\ref{sec:closed-form},
and writing $\alpha_i=m_i/\tau_i$, the standard normal integrals
\begin{equation}
\int_{-\alpha_i}^{\infty}\epsilon\phi(\epsilon)\,d\epsilon
=\phi(\alpha_i),
\qquad
\int_{-\alpha_i}^{\infty}\epsilon^2\phi(\epsilon)\,d\epsilon
=\Phi(\alpha_i)-\alpha_i\phi(\alpha_i)
\end{equation}
yield the mean $\mu_i^{+}$ and variance $v_i^{+}$ of
$a_i=\max(0,z_i)$ in
Eqs.~\eqref{eq:rectified-mean}--\eqref{eq:rectified-variance}.
For independent densities and fixed kernels,
\begin{equation}
\mathrm{Cov}_{q}\!\left(V(\mathbf{x}),V(\mathbf{x}')\right)
=\sum_i g_i(\mathbf{x})g_i(\mathbf{x}')v_i^{+}.
\end{equation}
Setting $\mathbf{x}'=\mathbf{x}$ recovers the predictive variance
in Eq.~\eqref{eq}.
Independent weights therefore do not imply independent voxels.
Exactness concerns these conditional moments under a fixed linear
operator; it does not establish an exact posterior over geometry
or a calibrated map of realized reconstruction error.

\paragraph{Sampling and implementation.}
The three frozen models in Section~\ref{sec:exp-closed-form} contain
83,048, 65,106, and 53,304 primitives for chest, foot, and beetle.
MC seeds 0--4 define independent sampling streams; budgets within
each stream are nested prefixes. Production voxelization uses
float32, with online float64 accumulation of MC means and variances.
The separate float64 check uses 512 foreground coordinates and
512 coordinates from its complement per scene, retaining fixed
spatial support while disabling density-dependent response pruning.
The complement of foreground includes transition voxels.

Table~\ref{tab:app-mc-ranking} connects numerical agreement to error
localization: high-budget sampling reproduces the limited foreground
ranking of chest and foot. All rankings target the same reconstruction
error $e(\mathbf{x})$. Agreement is not uniform across regions:
beetle retains a whole-volume MC-minus-CF Spearman difference of
$+0.0928$. Floating-point arithmetic, support handling, and production
pruning of contributions below $10^{-6}$ can affect the implemented
operator; finite MC comparisons do not prove implementation equality.

\begin{table}[htbp]
\centering
\caption{Foreground checks at $S=512$. Relative $L_2$ compares MC and
CF predictive variance maps, with the CF norm as denominator.
Here $\rho$ denotes Spearman correlation between predictive standard
deviation and reconstruction error $e(\mathbf{x})$. MC entries are medians over five sampling streams
for one frozen reconstruction per scene.}
\label{tab:app-mc-ranking}
\begin{tabular}{@{}lrrr@{}}
\toprule
Scene & Variance rel. $L_2$ (\%) & CF $\rho$ & MC $\rho$ \\
\midrule
chest  & 7.52 & 0.0810 & 0.0804 \\
foot   & 7.66 & 0.1774 & 0.1772 \\
beetle & 8.68 & 0.4547 & 0.4542 \\
\bottomrule
\end{tabular}
\end{table}

Timing uses an NVIDIA GeForce RTX 4060 Laptop GPU. The speedup in
Table~\ref{tab:closed-form-mc} is the median across scenes of the
ratio of median MC time to median CF time. Each median uses five
measured runs after two warm-ups. These timings cover complete
$256^3$ moment readout, including MC sampling and accumulation,
not the full reconstruction and error-localization system.

\section{Controls for the Foreground-Ranking Deficit}
\label{app:regions}

\paragraph{Threshold sensitivity.}
Table~\ref{tab:app-thresholds} tests whether the finding depends on
choosing 5\% of the reference peak. Within each uncertainty construction,
the reconstruction and score are unchanged when the region changes.
For the rectified-Gaussian model, the archived evaluator samples a
fixed random 5\% of each region's voxels, computes Spearman correlation
per reconstruction, and averages the three reconstruction seeds within
each scene. The deterministic ensemble uses five members and the error of their mean.

\begin{table}[htbp]
\centering
\caption{Median Spearman correlation across 15 scenes at 25 views.
Foreground contains voxels strictly above the stated percentage of
$V_{\max}^{*}$. Rectified-Gaussian entries first average three reconstruction seeds.
These intensity masks are not organ segmentations.}
\label{tab:app-thresholds}
\begin{tabular}{@{}lrr@{}}
\toprule
Region & Rectified-Gaussian density & Deterministic ensemble \\
\midrule
Whole volume & 0.846 & 0.849 \\
Foreground, 1\% & 0.231 & 0.359 \\
Foreground, 2\% & 0.189 & 0.267 \\
Foreground, 5\% & 0.108 & 0.199 \\
Foreground, 10\% & 0.104 & 0.164 \\
\bottomrule
\end{tabular}
\end{table}

\paragraph{Reconstruction-defined foreground and interior.}
The additional control in Section~\ref{sec:exp-diagnosis} fixes a
mask from the returned reconstruction at reconstruction seed 0. Its threshold is 5\%
of the 99.99th percentile of strictly positive reconstruction values;
all three reconstruction seeds share this mask. Interior distance is measured to
the nearest outside voxel center after one-voxel outside padding,
with strata $(0,2]$, $(2,5]$, and $>5$ grid voxels. Up to 100,000
fixed voxels are sampled per region. These distances are in grid
units, not anatomical thicknesses in millimeters.

This control targets $|\mu_q-V^{*}|$, rather than $|\widehat{V}-V^{*}|$,
and reports scene means after averaging reconstruction seeds. The foreground
correlation is 0.146 with a 95\% scene-bootstrap interval
$[0.091,0.206]$, and the $>5$-voxel interior correlation is 0.197
$[0.131,0.270]$. Masks do not use the reference, although the inherited
simulation and volume preprocessing do. These results test mask
choice and boundary effects, without identifying a unique cause
of weak error localization.

\paragraph{Shared-error diagnostics.}
The covariance analysis uses scores from reconstruction seed 0 and a fixed 10\% voxel
sample, with region weights given by sampled voxel proportions. It decomposes
raw covariance, not Spearman correlation. The finite-member squared-error
decomposition uses population variance with denominator $K$ and
float64 accumulation over the evaluated regions. The separate
correlations with member variation and shared absolute error use
a fixed 5\% voxel sample. Neither a large shared fraction nor these
associations alone establish a causal explanation of ranking failure.

\section{Learned Error Predictor: Implementation and Ablations}
\label{app:learner}

\subsection{Complete feature specification}
\label{app:learner-features}
Define the feature-normalization mask
$\mathcal{F}_{\mathrm{rec}}=
\{\widehat{V}>0.05Q_{99.99}(\widehat{V})\}$,
where $Q_p$ denotes the $p$th percentile, computed over the whole
volume unless a mask is specified. This mask differs
from the reference foreground $\mathcal{F}$ and from the positive-value
percentile mask used in Appendix~\ref{app:regions}.
Let $c_V=Q_{99}(\widehat{V}|\mathcal{F}_{\mathrm{rec}})+10^{-12}$ and
$c_\sigma=Q_{99}(\sigma_q|\mathcal{F}_{\mathrm{rec}})+10^{-12}$.
The normalized reconstruction and predictive standard deviation are
$\widetilde{V}=\widehat{V}/c_V$ and
$\widetilde{\sigma}_q=\sigma_q/c_\sigma$.
Local statistics use cubic windows with radius $\ell$ and side
$2\ell+1$ voxels. Derivatives use Gaussian filters with the stated
standard deviation in voxels and reflected boundaries.

\begin{table}[htbp]
\centering
\caption{Complete IMG and SIG features. Image derivatives use
$\widetilde{V}$; dispersion derivatives use $\widetilde{\sigma}_q$.
Counts sum to 35 and six. DIS and RES are specified in the text.}
\label{tab:app-features}
\begin{tabular}{@{}lp{0.73\linewidth}r@{}}
\toprule
Group & Descriptors & Count \\
\midrule
IMG & Normalized reconstruction $\widetilde{V}$ & 1 \\
& Local mean and standard deviation, $\ell=1,2,4,8$ & 8 \\
& Signed high-pass residual
$\widetilde{V}-\mathrm{mean}_{\ell}(\widetilde{V})$,
$\ell=1,2,4$ & 3 \\
& Absolute high-pass residual, $\ell=1,2,4,8$ & 4 \\
& Gradient magnitude at Gaussian scales 1, 2, and 4 & 3 \\
& Absolute gradient components at scale 1 & 3 \\
& Gradient anisotropy
$(\gamma_{\max}-\gamma_{\min})/(\gamma_{\max}+\gamma_{\min}+10^{-8})$,
where $\gamma_{\max}$ and $\gamma_{\min}$ are the largest and smallest
absolute gradient components & 1 \\
& Signed and absolute Gaussian Laplacian at scale 1 & 2 \\
& Local intensity range, $\ell=2$ & 1 \\
& In-plane radius and absolute axial offset & 2 \\
& Boundary distances inside and outside $\mathcal{F}_{\mathrm{rec}}$ & 2 \\
& Distance to strong edges and local mean gradient magnitude,
$\ell=4$ & 2 \\
& Absolute radial and tangential gradients and tangential fraction & 3 \\
\midrule
SIG & $\widetilde{\sigma}_q$ and
$\sigma_q/(\widehat{V}+10^{-3}c_V)$ & 2 \\
& Local mean of $\widetilde{\sigma}_q$, $\ell=2,4$ & 2 \\
& Local standard deviation of $\widetilde{\sigma}_q$, $\ell=2$ & 1 \\
& Gradient magnitude of $\widetilde{\sigma}_q$ at scale 1 & 1 \\
\bottomrule
\end{tabular}
\end{table}

Let $\widetilde{\mathbf{x}}=(\widetilde{x}_1,\widetilde{x}_2,
\widetilde{x}_3)$ denote coordinates normalized to $[-1,1]$,
with the third coordinate along the rotation axis. In-plane radius
and absolute axial offset are
$\sqrt{\widetilde{x}_1^2+\widetilde{x}_2^2}$ and
$|\widetilde{x}_3|$. Boundary distances are divided by the largest
volume dimension. Strong edges exceed the 80th percentile of
scale-1 gradient magnitude within $\mathcal{F}_{\mathrm{rec}}$;
edge distance is clipped at 32 voxels and divided by 32.
Let $(\nu_1,\nu_2)=(\widetilde{x}_1,\widetilde{x}_2)/
(\sqrt{\widetilde{x}_1^2+\widetilde{x}_2^2}+10^{-6})$ and let
$(\gamma_1,\gamma_2,\gamma_3)$ be the scale-1 derivatives of
$\widetilde{V}$ along the three axes. The directional features are
$\gamma_{\mathrm{rad}}=|\nu_1\gamma_1+\nu_2\gamma_2|$,
$\gamma_{\mathrm{tan}}=|-\nu_2\gamma_1+\nu_1\gamma_2|$, and
$\gamma_{\mathrm{tan}}/
(\gamma_{\mathrm{rad}}+\gamma_{\mathrm{tan}}+10^{-8})$.

For each classical reconstruction $\widehat{V}^{\mathrm{cl}}_j$,
DIS contains $d_j=(\widehat{V}-\widehat{V}^{\mathrm{cl}}_j)/c_V$,
$|d_j|$, and the $5^3$ local mean of $|d_j|$.
The index $j$ enumerates FDK, SART, and ASD-POCS, using the same
25 views. Two additional features are the population standard
deviation across $\widehat{V}^{\mathrm{cl}}_j/c_V$ and
$|\widehat{V}-\mathrm{median}_j(\widehat{V}^{\mathrm{cl}}_j)|/c_V$,
giving 11 DIS features. For the projection residual
$r=A\widehat{V}-y$, the three RES features are $|A^\top(|r|)|$,
$|\mathrm{FDK}(r)|$, and a $5^3$ local mean of the latter.
Each of the first two maps is divided by its whole-volume 99th
percentile plus $10^{-12}$ before local averaging.

\subsection{Sampling, targets, and statistical aggregation}
The common feature cache contains up to 150,000 indices per scene,
sampled without replacement from the union of the reference foreground
and the three reconstruction foregrounds. All arms use the same rows;
evaluation restricts them to $\mathcal{F}$. Each leave-one-scene-out
fold draws 80,000 foreground training rows per remaining scene from
reconstruction seed 0, with row seed 0. Feature values require no test
reference, but this archived evaluation-row selection uses reference
data.

For each training scene, target normalization precedes the 80,000-row
subsampling. Using its cached foreground rows, Shared supervision sets
$\delta=\max(10^{-3}\mathrm{median}(e_{\mathrm{shared}}),10^{-12})$,
$\mu_{\log}=\mathrm{mean}(\log(e_{\mathrm{shared}}+\delta))$, and
$\sigma_{\log}=\mathrm{std}(\log(e_{\mathrm{shared}}+\delta))+10^{-12}$,
using population standard deviation. The training target
$t(\mathbf{x})$ is then given by
Eq.~\eqref{eq:predictor-training-target}. Individual supervision
replaces $e_{\mathrm{shared}}$ with the training reconstruction's
error $e$ in these definitions. No held-out reference statistics
enter prediction.
In addition to the capacity settings in Section~\ref{sec:exp-repair},
LightGBM uses squared-error regression, at least 100 rows per leaf,
row subsampling 0.8 each round, feature subsampling 0.7, $L_2$ penalty
2, four CPU threads, and deterministic column-wise histograms.

The four main arms use learner seeds 0--2 and reconstruction seeds
0--2: 180 fitted models and 540 evaluations. Metrics first average
reconstruction seeds within scene and learner seed, then learner seeds
within scene, then the 15 scenes equally. This is metric averaging,
not prediction ensembling. Error scores $s(\mathbf{x})$ are ranked directly without
exponentiation. Normalized RMSE AUSE uses 20 equally spaced removal fractions from
0 to 0.95 and trapezoidal integration, normalized by initial
foreground RMSE. Paired 95\% intervals use 20,000 scene bootstrap
resamples with seed 20260913; neither seeds nor voxels are independent
scene replicates. Intervals are conditional on this benchmark and
its overlapping training folds; secondary comparisons are unadjusted.

\begin{table}[htbp]
\centering
\caption{Paired effects after averaging both seed types within each
scene. Differences are first minus second; positive $\Delta\rho$ and
negative $\Delta\mathrm{AUSE}$ favor the first configuration. Brackets give
95\% scene-bootstrap intervals. Only the first comparison is primary.}
\label{tab:app-contrasts}
\begin{tabular}{@{}>{\raggedright\arraybackslash}p{0.32\linewidth}p{0.29\linewidth}p{0.29\linewidth}@{}}
\toprule
Comparison & $\Delta\rho$ [95\% interval] & $\Delta\mathrm{AUSE}$ [95\% interval] \\
\midrule
Full/Shared vs.\ IMG/Individual
& $+0.0683$\newline $[+0.0473,+0.0937]$
& $-0.0720$\newline $[-0.0953,-0.0509]$ \\
Full/Shared vs.\ Full/Individual
& $+0.0014$\newline $[-0.0060,+0.0082]$
& $-0.0057$\newline $[-0.0104,-0.0008]$ \\
Full/Shared vs.\ IMG+DIS+RES/Shared
& $+0.0086$\newline $[+0.0010,+0.0147]$
& $-0.0116$\newline $[-0.0168,-0.0060]$ \\
\bottomrule
\end{tabular}
\end{table}

The Full/Shared versus IMG/Individual mean $\Delta\rho$ is 0.0689,
0.0693, and 0.0666 for learner seeds 0, 1, and 2, respectively.
With all 55 input features fixed, the Spearman correlation interval
for Shared minus Individual supervision spans
zero. The SIG increment is positive in 13/15 scene averages, with
exceptions for jaw and backpack. Thus, the results support complementary
inputs, without attributing the joint gain to shared supervision alone.

\subsection{Fixed-target feature ablations}
Table~\ref{tab:app-ablation} isolates input changes under Shared
supervision. These comparisons use learner seed 0, averaging three
reconstruction-seed metrics within each scene before averaging scenes.
The untrained ASD-POCS discrepancy score is the fixed $5^3$ local
absolute difference from the ASD-POCS reconstruction. The untrained
residual backprojection score is normalized $|A^\top(|r|)|$. Neither requires fitting an error predictor.

\begin{table}[htbp]
\centering
\caption{Single-learner-seed ablations on the 15-scene benchmark.
All learned rows use Shared supervision. Raw dispersion denotes
$\sigma_q$. Values are scene means of
foreground Spearman correlation and normalized RMSE AUSE. Full
combines all 55 features. These estimates are separate from the
three-learner-seed results in Table~\ref{tab:matched-learner}.}
\label{tab:app-ablation}
\begin{tabular}{@{}lrr@{}}
\toprule
Inputs or score & Foreground $\rho$ & AUSE \\
\midrule
IMG & 0.3399 & 0.3683 \\
IMG+DIS & 0.3668 & 0.3379 \\
IMG+RES & 0.3726 & 0.3284 \\
IMG+DIS+RES & 0.3901 & 0.3156 \\
Full & 0.3994 & 0.3039 \\
SIG & 0.1951 & 0.4529 \\
\midrule
Raw dispersion & 0.1357 & 0.5417 \\
ASD-POCS discrepancy & 0.2654 & 0.3875 \\
Residual backprojection & 0.1397 & 0.4788 \\
\bottomrule
\end{tabular}
\end{table}

Equal learner budgets do not imply equal total costs: Shared labels
require three training reconstructions, DIS requires auxiliary
classical reconstructions, and RES requires projection and
backprojection. The experiments establish predictive value of these
inputs, not a unique identification of shared error or an end-to-end
speed advantage.

\section{Transfer with the Earlier Predictor}
\label{app:transfer}

The transfer study referenced in Section~\ref{sec:exp-repair}
uses an earlier 52-feature error predictor (IMG+SIG+DIS, without RES)
and Shared supervision. It does not evaluate the current 55-feature
configuration. Training uses 1,200 boosting rounds, 255 leaves,
learning rate 0.04, and up to 80,000 foreground rows per training
scene from reconstruction seed 0.

\begin{table}[htbp]
\centering
\caption{Earlier-predictor transfer. Values are median foreground
Spearman correlations across scenes or specimens, after averaging
available reconstruction seeds within each case. NAF uses ensemble
standard deviation as its dispersion baseline.}
\label{tab:app-transfer}
\begin{tabular}{@{}lrrr@{}}
\toprule
Evaluation & Cases & Predictor & Dispersion \\
\midrule
50-view benchmark & 15 & 0.338 & 0.102 \\
75-view benchmark & 15 & 0.333 & 0.105 \\
Neural attenuation field (NAF) & 3 & 0.461 & 0.168 \\
Measured specimens & 3 & 0.421 & 0.191 \\
\bottomrule
\end{tabular}
\end{table}

For view transfer, each predictor is trained on the other 14 scenes
at 25 views and applied without refitting to its held-out scene.
Each higher budget has eight scenes with three reconstruction seeds
and seven with one, totaling 31 reconstructions. DIS retains the
same scene's 25-view classical reconstructions.
For NAF, all three target scenes---chest, bonsai, and broccoli---are
excluded from training, leaving 12 benchmark scenes.
The target inputs and errors use the NAF ensemble mean, with its
ensemble standard deviation as dispersion. For pine, seashell, and
walnut, a separate predictor trains on all 15 benchmark scenes and
uses 25 measured views per specimen. Spearman correlations are
averaged over three reconstruction seeds per specimen; the reference is a 721-view FDK reconstruction,
so evaluation inherits its reconstruction errors.

\section{Human-Anatomy Experimental Details}
\label{app:human}

\paragraph{Data and preprocessing.}
The ten additional training subjects are MSD lung cases 001, 003,
and 004; pancreas 001, 004, and 005; spleen 10 and 12; and colon 001
and 005. The original test set uses lung 005, pancreas 006, spleen 13,
colon 006, and the first four lexicographically sorted LUNA16
\texttt{subset0} series. Selection was recorded before image download.
The additional test cohort contains four further MSD and four further
LUNA16 subjects. Both test cohorts are disjoint from predictor training.

The original acquisition protocol orients volumes to LPS, clips
intensities to $[-1000,3000]$ HU, normalizes by
$(\mathrm{HU}+1000)/4000$, and resamples to $256^3$. Body cropping
uses a $-700$ HU threshold with an 8\% margin. Simulated 25-view
projections span $360^{\circ}$ using the chest cone-beam geometry;
stored noise parameters are Poisson $10^5$ and additive Gaussian
$(0,10)$. These are simulated acquisitions, not measured patient scans.

\paragraph{Models and fusion.}
The human-anatomy tree predictor uses the 55 features, at most
3.6 million training rows across reconstruction seeds 0--2, and the
same tree-capacity settings as the predictor in
Table~\ref{tab:matched-learner}. Anatomy-augmented training assigns equal
row budgets to benchmark and human data. The 3D U-Net uses eight image-derived
channels: $\widetilde{V}$, $\widetilde{\sigma}_q$, the three $|d_j|$
maps, normalized $|\mathrm{FDK}(r)|$, the standard deviation of
normalized classical reconstructions, and a $9^3$ local mean of
$\widetilde{\sigma}_q$. Two further channels encode
normalized in-plane radius and absolute axial offset.

The 3D U-Net uses base width 32, $64^3$ patches, batch size six, and
8,000 Adam updates with initial learning rate $3\times10^{-4}$,
cosine decay, and exponential-moving-average decay 0.999. Training
minimizes foreground absolute loss on the standardized log
shared-error target $t(\mathbf{x})$, with target statistics computed
over each training volume's dense foreground. Anatomy-augmented batches contain three benchmark and three
human patches. Inference averages four in-plane flips and overlapping
patches with stride 48. Component scores are converted to ranks over
all 150,000 stored rows, with average ranks for ties, and averaged
equally before foreground evaluation. As in Table~\ref{tab:anatomy},
Anatomy and Benchmark denote the rank-fusion predictors trained
with and without the ten additional human training subjects.

\paragraph{Evaluation scope.}
The original three-reconstruction-seed check changes the tree
predictor inputs and reconstruction error $e(\mathbf{x})$ but reuses
the stored 3D U-Net error scores
across seeds; metrics are then averaged within subject. It is not
three independent 3D U-Net inferences. The additional eight subjects
were reconstructed again after historical outcomes were known,
using frozen predictor checkpoints and fusion weights; their results
are retrospective. One inspected additional-cohort checkpoint has
nonfinite raw parameters for Gaussian spatial scales despite finite
activated spatial scales and evaluated outputs; its effect on ranking has not been isolated.
The human study supports error-ranking comparisons under these
protocols, without establishing calibrated error magnitudes or
clinical diagnostic benefit.